\pdfoutput=1

\documentclass[11pt]{article}

\usepackage[preprint]{acl}

\usepackage{times}
\usepackage{latexsym}

\usepackage[T1]{fontenc}

\usepackage[utf8]{inputenc}

\usepackage{microtype}

\usepackage{inconsolata}

\usepackage{graphicx}

\usepackage{amsmath}

\usepackage{booktabs}

\usepackage{pifont}
\newcommand{\cmark}{\ding{51}}
\newcommand{\xmark}{\ding{55}}
\usepackage{amsfonts}
\usepackage{algorithm}
\usepackage{algpseudocode}
\usepackage{placeins}
\usepackage{float}
\usepackage{tabularx}
\usepackage{multirow}
\usepackage{enumitem}

\makeatletter
\renewcommand{\paragraph}{%
  \@startsection{paragraph}{4}%
  {\z@}{0.5ex \@plus 1ex \@minus .2ex}{-1em}%
  {\normalfont\normalsize\bfseries}%
}
\makeatother

\title{PHLoRA: data-free Post‑hoc Low‑Rank Adapter extraction from full-rank checkpoint}

\author{
  Bhoomit Vasani \\ Amazon AGI \\ \texttt{vbhoomit@amazon.com} \And
  Jack FitzGerald \\ EdgeRunner AI \\ \texttt{jack@edgerunnerai.com} \And
  Anjie Fang \\ Amazon AGI \\ \texttt{njfn@amazon.com}
  \AND
  Sushmit Vaish \\ Amazon AGI \\ \texttt{sushvai@amazon.com}
}

\begin{document}
\maketitle
\begin{abstract}
We introduce \textbf{PHLoRA}\footnote{Pronounced ``flora''.} (Post-hoc LoRA), a simple yet powerful method to extract low-rank adaptation adapters from full-rank fine-tuned models without requiring access to training data or gradients. By computing the low-rank decomposition of weight differences between a base model and its fine-tuned counterpart, our method reconstructs adapter modules that can be merged or dynamically routed at inference time via S-LoRA\citep{bing2024slora}, or served in scalable, industry settings using platforms like NVIDIA NIM\footnote{\href{https://developer.nvidia.com/blog/seamlessly-deploying-a-swarm-of-lora-adapters-with-nvidia-nim/}{NVIDIA NIM deployment blog}}.
This approach amortizes latency overhead across requests and yields substantial cost savings. Unlike prior work that trains each adapter explicitly, our approach decouples fine-tuning from adapter generation, allowing adapter extraction from existing full-rank models or third-party checkpoints. Experiments on text, image, and video benchmarks using the Amazon Nova~\citep{novatechreport} model family demonstrate that extracted adapters preserve high energy from the full weight delta, can be pruned safely, and yield negligible degradation in downstream task performance when re-merged. Overall, PHLoRA provides a practical path for making all existing full-rank checkpoints adapter-ready, democratizing scalable inference for all models.
\end{abstract}

\section{Introduction}
The Low‑Rank Adapters (LoRA) technique~\citep{hu2022lora} is a popular way to reduce memory during training, and it offers an additional advantage at inference: it allows a single server to host adapters for hundreds or thousands of users in a shared inference API, as in S‑LoRA~\citep{bing2024slora}. Modern industry platforms such as NVIDIA NIM\footnotemark[2] support scalable, low-latency serving of LoRA-based adapters in production. However, many practitioners have existing models trained with full‑rank fine‑tuning, including through the use of other training methods beyond standard fine-tuning like DPO~\citep{rafailov2024directpreferenceoptimizationlanguage} or PPO~\citep{Schulman2017ProximalPO}. To serve these users, we introduce and evaluate a method for compressing full‑rank updates into low‑rank adapters compatible with dynamic serving frameworks, called \textbf{Post‑hoc Low‑Rank Adapter Extraction} (PHLoRA).
Our contributions include the following:
\begin{itemize}
\itemsep0em
\item \textbf{Post‑hoc LoRA formulation:} We pose adapter extraction as a low‑rank decomposition solved with truncated SVD over the checkpoint’s weight delta, requiring a single forward pass and no gradients or data.

\item \textbf{Flexible deployment and fast start‑up:} Compact adapters cut model‑load latency by over 10$\times$ compared to full‑rank checkpoints and can be merged for static inference or dynamically routed via shared-adapter execution (e.g., S-LoRA), and are compatible with scalable industry platforms such as NVIDIA NIM, to minimize run-time cost.
\item \textbf{Multimodal results:} We evaluate on three text, one image, and one video understanding benchmark, showing PHLoRA preserves performance while reducing inference cost by up to 4$\times$.
\end{itemize}
We provide all dataset processing code, modeling code, and evaluation prompts\footnote{github URL to be added}.

\section{Background and Related Work}

PHLoRA uniquely provides constant‑cost, post‑hoc adapter generation that is fully LoRA‑inference-compatible for both text and multimodal settings~\citep{sung2022vladapterparameterefficienttransferlearning}, with further comparisons in Table~\ref{tab:rw-compare}.

\begin{table*}[ht]
  \centering
  \small
  \setlength{\tabcolsep}{3pt}
  \begin{tabular}{lccccccc}
    \toprule

    \textbf{Method} & \textbf{Stage} & \textbf{Input} & \textbf{SVD On} & \textbf{Output} & \textbf{LoRA‑comp.} & \textbf{Task‑spec. Train?} & \textbf{Needs Data?} \\
    \midrule
    PHLoRA (ours) & Post-hoc & $\Delta W$ & $\Delta W$ & LoRA $A,B$ & \cmark & \xmark & \xmark \\
    SLiM & Post-hoc & $W$ & $W$ & LR+Q weights & \xmark & \xmark & \xmark \\
    SVD-LLM & Post-hoc & $W$ & $W$ & Trunc. LR model & \xmark & \xmark & \xmark \\
    SVDQuant & Post-hoc & $W$ & $W$ & LR+Q weights & \xmark & \xmark & \xmark \\
    Dobi-SVD & Post-hoc+Grad & $W$ & Diff. SVD & Compressed model & \xmark & \cmark & \cmark \\
    SORSA & PEFT init & $W$ & $W$ & Struct. adapter & $\triangle$ & \cmark & \cmark \\
    PiSSA & PEFT init & $W$ & $W$ & Init. adapter & \cmark & \cmark & \cmark \\

    \bottomrule
  \end{tabular}
  \caption{Qualitative comparison of PHLoRA and related approaches. \cmark: yes; \xmark: no; $\triangle$: partially.}
  \label{tab:rw-compare}
\end{table*}

LoRA inserts rank‑$r$ matrices in parallel with linear layers and trains only these additions, reducing memory and compute~\citep{hu2022lora}. LoRA+ further re‑balances the optimizer by raising the learning rate on the $B$ matrix~\citep{hayou2024loraefficientlowrank}. Other variants explore dynamic rank schedules (AdaLoRA~\citep{zhang2023adalora}), quantized training (QLoRA~\citep{dettmers2023qlora}), and selective layer targeting.
Soft prompt‑tuning~\citep{lester2021power}, BitFit~\citep{zaken2021bitfit}, AdapterFusion~\citep{pfeiffer2021adapterfusion}, and VL-Adapter~\citep{sung2022vladapterparameterefficienttransferlearning} trade different portions of trainable parameters for efficiency, but all require task‑specific optimization. Recent methods extend parameter-efficient transfer to vision-language models~\citep{sung2022vladapterparameterefficienttransferlearning}.
PiSSA initializes LoRA adapters with principal singular vectors \emph{before} adapter training, accelerating convergence but not eliminating the need for optimization~\citep{meng2025pissaprincipalsingularvalues}.
SLiM~\citep{mozaffari2025slimoneshotquantizationsparsity}, SVD-LLM~\citep{li2023svdllm}, and SVDQuant~\citep{zhou2024svdquant} apply low-rank decomposition (often combined with quantization) directly to pretrained weights $W$ for inference compression and acceleration. GPTQ~\citep{frantar2023gptqaccurateposttrainingquantization} is another widely-used post-hoc quantization approach. However, these methods do not expose LoRA‑compatible factors nor leverage the fine-tuning delta.
Dobi‑SVD~\citep{wang2025dobisvddifferentiablesvdllm} makes SVD differentiable and tunes the factors with task supervision, achieving lower reconstruction error at the cost of additional gradient steps.
SORSA~\citep{yuan2023sorsa} proposes a structured low‑rank adaptation that replaces dense LoRA matrices but still requires full adapter training.

While prior works have explored low-rank approximation techniques for fine-tuning (e.g., \citealp{hu2022lora, zhang2023adalora}), we also found a recent GitHub implementation, LoRD \citep{gauthier2024lord}, that performs similar post-hoc low-rank extraction, though without an associated peer-reviewed manuscript.

\section{Methodology}

\begin{table*}[htbp]
  \centering
  \small
  \begin{tabularx}{\linewidth}{@{} l X @{}}
    \toprule
    \textbf{Dataset} & \textbf{Metric(s)} \\
    \midrule
    TAT--QA~\citep{zhu2021tatqa}                & Accuracy, EM \\
    MKFE~\citep{mkfe_hf}                 & Key Overlap, Value Overlap (\%) \\
    MedMCQA~\citep{pmlr-v174-pal22a}       & Accuracy, F1 \\
    VQA--RAD~\citep{lau2018rad}          & Avg.\ Normalized Similarity \\
    CaptionGen~\citep{chen2011collecting}        & ROUGE\_L, CIDEr \\
    \bottomrule
  \end{tabularx}
  \caption{Primary evaluation metrics and citations for each benchmark.}
  \label{tab:metrics}
\end{table*}

\subsection{Problem Setup}\label{sec:setup}
Given a pretrained model and a fine-tuned model, each consisting of weights, we define the weight delta as

\begin{equation}
\Delta W = W_{\text{ft}} - W_{\text{base}}, \ \textrm{where} \ W\in\mathbb{R}^{d \times k}
\end{equation}
Our objective is to approximate each $\Delta W$ with a rank-$r$ factorization in the LoRA form:
\begin{equation}
\Delta W \approx BA, \textrm{where}\ 
A\in\mathbb{R}^{r \times k},
B\in\mathbb{R}^{d \times r}
\end{equation}
Once $A$ and $B$ are obtained, they can be deployed as standard LoRA adapters (for dynamic or conditional routing) or merged back into the backbone via $W_{\text{base}} \leftarrow W_{\text{base}} + BA$.
This process is repeated for all target components (typically attention and MLP submodules).

\subsection{Post-hoc LoRA Extraction}\label{sec:extraction}
We perform a truncated singular value decomposition (SVD) on $\Delta W$:
\begin{equation}
\begin{gathered}
\label{eq:svd_def}
U \Sigma V^{\top} = \operatorname{SVD}(\Delta W), \textrm{where} \\
U \in \mathbb{R}^{d \times d}, \quad
\Sigma \in \mathbb{R}^{d \times k}, \quad
V \in \mathbb{R}^{k \times k}  
\end{gathered}
\end{equation}

The low-rank LoRA factorization is then:
\begin{equation}
\begin{gathered}
B = U_{[:,:r]} \Sigma_{[:r,:r]}^{\frac{1}{2}} \\
A = \Sigma_{[:r,:r]}^{\frac{1}{2}} V^{\top}_{[:r,:]} 
\end{gathered}
\label{eq:ab}
\end{equation}
where the first $r$ columns of $U$, the first $r$ rows of $V$, and the first $r$ rows and columns of $\Sigma$ are taken, and the $\frac{1}{2}$ exponent represents the element-wise square root.
This SVD-based decomposition ensures that $BA$ is the best rank-$r$ approximation of $\Delta W$~\citep{Eckart1936TheAO}.
All computations are performed independently for each target weight matrix (e.g., $q_{\text{proj}}$, $k_{\text{proj}}$, $mlp_{\text{fc1}}$).

Merged inference computes $W_{\text{base}} + BA$ once, fully restoring the original fine-tuned model up to truncation error (no runtime adapter overhead).
Dynamic routing, as in S-LoRA~\citep{bing2024slora}, loads $A$ and $B$ as lightweight adapters and activates them on demand, enabling low-cost serving of multiple adapters in a single process.

PHLoRA is compatible with the HuggingFace PEFT library~\citep{peft2023}, PyTorch LoRA implementations, and multi-adapter serving frameworks. No access to gradients or training data is needed, but only the base and fine-tuned checkpoints.

\subsection{Energy-Based Analysis}\label{sec:energy}
\begin{figure}[]
\centering
\includegraphics[width=1.0\linewidth]{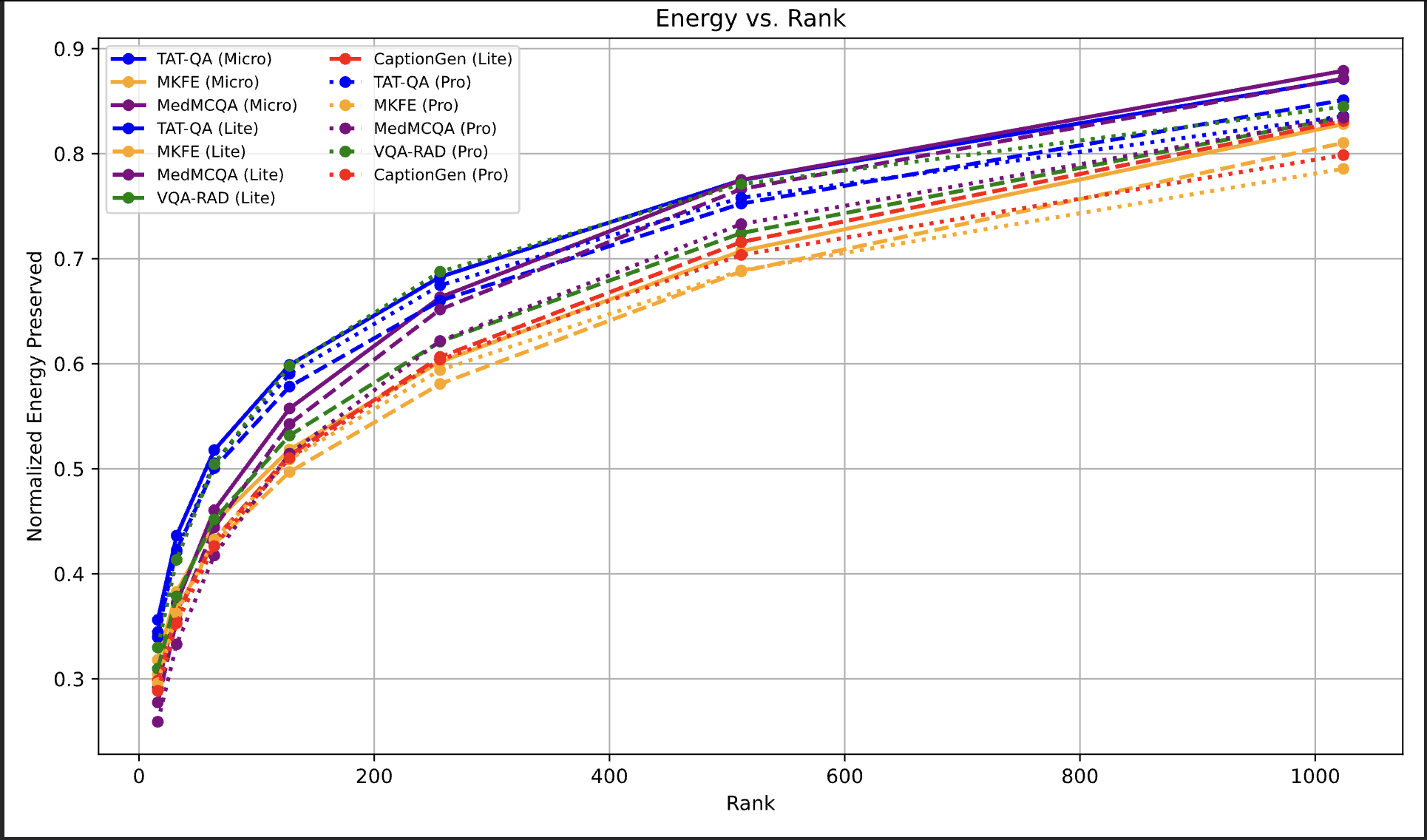}
\caption{Preserved energy vs rank.}
\label{fig:rank-ablation}
\end{figure}

In low-rank matrix approximation, the \textit{energy} of a matrix refers to the sum of the squares of its singular values, quantifying the total information content or signal present in the matrix. For a weight delta $\Delta W$ with singular values $\sigma_1, \sigma_2, \dots \sigma_d$, we define the preserved energy at rank $r$ as
\begin{equation}
E_r = \frac{\sum_{i=1}^{r} \sigma_i^{2}}{\sum_{i=1}^{d} \sigma_i^{2}}.
\label{eq:energy}
\end{equation}
Intuitively, $E_r$ measures what fraction of the ``important'' weight update is retained by the top-$r$ singular directions. High preserved energy typically correlates with the adapter's ability to recover full-rank performance. We report $E_r$ across all layers and multiple ranks (e.g., 32, 64), as visualized in Figure~\ref{fig:rank-ablation}. Although we fix $r$ globally in this work, our code supports per-layer adaptive rank selection based on a desired energy threshold.

\section{Experiments}

\subsection{Experimental Setup}
\label{sec:setup}
We benchmark \textbf{PHLoRA} on three text only datasets,
\textit{TAT‑QA}~\citep{zhu2021tatqa}, \textit{Medical Knowledge from Extracts (MKFE)}~\citep{mkfe_hf}, \textit{MedMCQA}~\citep{pmlr-v174-pal22a};
one image and text dataset, \textit{VQA‑RAD}~\citep{lau2018rad};
and one video and text dataset, \textit{CaptionGen}~\citep{chen2011collecting}.
We sub-sample and reformat the datasets.
See Table~\ref{tab:metrics} for primary metrics, and Appendix~\ref{app:datasets} for detailed statistics.
All experiments use the Nova model family. 

We compare: (i) the base model, (ii) the full‑rank fine‑tuned model, (iii) LoRA+ with rank~32, and (iv) PHLoRA with rank~32 (default) and 64 (see Section~\ref{sec:ablation} for larger $r$). We use an evaluation prompt that does not specify formatting, which results in many formatting errors in the base model. Though the base model could be improved substantially with prompt optimization, our choice of prompts accentuates the effect of fine-tuning.

\begin{table*}[htbp]
  \centering
  \small
  \caption{Test set results for Nova Micro (text-only), Lite, and Pro. Where two metrics are shown, the best per metric is \textbf{bolded}. The evaluation prompts do not provide formatting instructions, substantially increasing the difficulty of the task for the base model prior to fine-tuning.}
  \label{tab:all-results}
  \begin{tabularx}{\textwidth}{lcccccc}
    \toprule
    \textbf{Dataset (Metric)} & \textbf{Base Model} & \textbf{Full-Rank} & \textbf{LoRA+ (r32)} & \textbf{PHLoRA (r32)} & \textbf{PHLoRA (r64)} \\
    \midrule
    \multicolumn{6}{l}{\textit{Nova Micro}} \\
    \midrule
    TAT--QA (Acc / Exact Match)    & 0 / 0 & \textbf{84.95} / \textbf{51.68} & 82.47 / 48.49 & 83.54 / 48.14 & 84.07 / 49.02 \\
    MKFE (Key Overlap / Val Overlap) & 50.0 / 22.0  & \textbf{100.0} / \textbf {28.0} & \textbf{100.0} / 26.75  & 99.0 / 27.75 & 99.0 / \textbf {28.0} \\
    MedMCQA (Accuracy / F1)             & 0.03 / 0.05 & 60.49  /  60.52  & 60.52 / 60.52 & \textbf {60.79}  /  \textbf {60.85} & 60.60 / 60.63  \\
    \midrule
    \multicolumn{6}{l}{\textit{Nova Lite}} \\
    \midrule
    TAT--QA (Acc / Exact Match)      & 0 / 0 & 83.89 / 48.85 & 82.48 / 48.50 & 85.84 / 52.39 & \textbf{86.02} / \textbf{53.45} \\
    MKFE (Key Overlap / Val Overlap)    & 49.50 / 19.50        & \textbf{99.50} / 22.75       & 99.0 / 24.75  & \textbf{99.50} / \textbf{26.50} & \textbf{99.50} / 26.0  \\
    MedMCQA (Accuracy / F1)               & 0.19 / 0.38 & 63.40 / 63.33 & 59.52 / 59.52 & 64.11 / 64.07 & \textbf{64.30} / \textbf{64.25}  \\
    VQA--RAD (Avg Norm Similarity)  & 23.22        & 54.87     & 57.15        & \textbf{58.56}        & 57.56 \\
    CaptionGen (ROUGE\_L / CIDEr)         & 31.37 / 0.81  & 49.40 / 1.43    & 48.26 / 1.46  & 49.19 / \textbf{1.51} & \textbf{49.43} / 1.50 \\
    \midrule
    \multicolumn{6}{l}{\textit{Nova Pro}} \\
    \midrule
    TAT--QA (Acc / Exact Match)      & 0 / 0 & \textbf{89.38} / \textbf{62.48} & 87.79 / 53.98 & 87.96 / 55.22 & 89.00 / 58.00 \\
    MKFE (Key Overlap / Val Overlap)    & 50.0 / 17.0        & 99.50 / 24.75        & 99.50 / \textbf{25.50}        & \textbf{100.0} / 24.0        & \textbf{100.0} / 25.0 \\
    MedMCQA (Accuracy / F1)               & 0 / 0 & 69.40 / 69.42 & \textbf{71.14} / \textbf{71.16} & 70.0 / 70.0 & 70.0 / 70.0 \\
    VQA--RAD (Avg Norm Similarity)  & 27.03        & 56.20     & 56.57        & \textbf{57.58}        & 56.92 \\
    CaptionGen (ROUGE\_L / CIDEr)         & 37.63 / 1.13  & 48.94 / 1.37  & \textbf{50.12} / \textbf{1.55} & 48.55 / 1.45 & 48.85 / 1.48 \\
    \bottomrule
  \end{tabularx}
\end{table*}

\subsection{Results and Analysis}
\label{sec:results}

Table~\ref{tab:all-results} reports test-set performance across all model sizes and benchmarks. For each task, we include one or two evaluation metrics (e.g., Accuracy / Exact Match), with the best score for each metric shown in \textbf{bold}. PHLoRA demonstrates consistency with full-rank fine-tuning across the Nova Micro, Lite, and Pro model families, often coming within 1\% performance while occasionally even surpassing full-rank results.

On Nova Micro, full-rank leads on TAT-QA, but PHLoRA remains close and even surpasses it on MedMCQA, with only minor gaps on MKFE. On Nova Lite, PHLoRA (r64) delivers the best scores on TAT-QA, MedMCQA, VQA-RAD, and CaptionGen, which highlights its strength in reasoning and multimodal tasks. Nova Pro further demonstrates scalability: PHLoRA nearly matches full-rank on TAT-QA and outperforms it on MedMCQA and VQA-RAD, while it also remains competitive on CaptionGen. Overall, the margin between PHLoRA and Full-Rank shrinks as Nova model scales, with PHLoRA often taking the lead.

\paragraph{Inference Cost and Latency.}
PHLoRA, when merged into the backbone (“m-packed” as in S-LoRA~\citep{bing2024slora}), is computationally equivalent to full-rank and merged LoRA inference for a single adapter or task. All three approaches require only a single matrix multiplication per layer. For scalable multi-adapter deployment, we estimate cost and throughput improvements using S-LoRA-like dynamic routing~\citep{bing2024slora}, which achieves up to 4$\times$ higher throughput and cost efficiency than naive dynamic LoRA serving (e.g., PEFT or vLLM) in multi-tenant settings, as shown in Table~3 and Figure~4 of S-LoRA. These reference results provide a strong indication that PHLoRA, when paired with S-LoRA-like serving, is highly cost-effective for scalable, multi-user inference scenarios.\footnote{We use “S-LoRA-like” to refer to any scalable, dynamic multi-adapter LoRA serving implementation; S-LoRA~\citep{bing2024slora} is used as a reference.}

\begin{table*}[h]
  \centering
  \small
  \caption{Ablation study for Nova Micro (text-only). We show one or two evaluation metrics, with the best score for each metric shown in \textbf{bold} and $E_r$ (preserved energy, \%) for PHLoRA in parentheses.}
  \label{tab:ablation-micro}
  \begin{tabularx}{\textwidth}{lccccc}
    \toprule
    \textbf{Dataset (Metric)} & \textbf{Full-Rank} & \textbf{PHLoRA (r32)} & \textbf{PHLoRA (r64)} & \textbf{PHLoRA (r512)}  \\
    \midrule
    TAT--QA (Accuracy / EM)      & \textbf{84.96} / \textbf{51.68} & 83.54 / 48.14 (44) & 84.07 / 49.03 (52) & \textbf{84.96} / 51.15 (77) \\
    MKFE (Key Overlap / Value Overlap) & \textbf{100.0} / 28.0 & 99.0 / 27.75 (38) & 99.0 / 28.0 (45) & \textbf{100.0} / \textbf{28.75} (71) \\
    MedMCQA (Accuracy / F1)      & 60.49 / 60.52 & 60.79 / 60.85 (37) & 60.60 / 60.63 (46) & \textbf{60.93} / \textbf{60.94} (78) \\
    \bottomrule
  \end{tabularx}
\end{table*}

\begin{table*}[h]
  \centering
  \small
  \caption{Ablation study for Nova Lite (text, image, video). We show one or two evaluation metrics, with the best score for each metric shown in \textbf{bold} and $E_r$ (preserved energy, \%) for PHLoRA in parentheses.}
  \label{tab:ablation-lite}
  \begin{tabularx}{\textwidth}{lccccc}
    \toprule
    \textbf{Dataset (Metric)} & \textbf{Full-Rank} & \textbf{PHLoRA (r32)} & \textbf{PHLoRA (r64)} & \textbf{PHLoRA (r512)} \\
    \midrule
    TAT--QA (Accuracy / EM)       & 83.89 / 48.45 & 85.84 / 52.39 (42) & 86.02 / 53.45 (49) & \textbf{86.90} / \textbf{55.58} (74) \\
    MKFE (Key Overlap / Value Overlap) & \textbf{99.50} / 22.75 & \textbf{99.50} / \textbf{26.50} (36) & \textbf{99.50} / 26.00 (42) & 99.0 / 24.75 (68) \\
    MedMCQA (Accuracy / F1)       & 63.40 / 63.33 & 64.11 / 64.07 (35) & \textbf{64.30} / \textbf{64.25} (44) & 63.83 / 63.77 (76) \\
    VQA--RAD (Similarity)           & 54.87        & 58.57 (37)        & \textbf{57.56} (45)                 & 55.01 (71)        \\
    CaptionGen (ROUGE\_L / CIDEr)  & 49.40 / 1.43        & 49.19 / \textbf{1.51} (36) & 49.43 / 1.50 (43)                 & \textbf{49.84} / 1.50 (71) \\
    \bottomrule
  \end{tabularx}
\end{table*}

\begin{table*}[h]
  \centering
  \small
  \caption{Ablation study for Nova Pro (text, image, video). We show one or two evaluation metrics, with the best score for each metric shown in \textbf{bold} and $E_r$ (preserved energy, \%) for PHLoRA in parentheses.}
  \label{tab:ablation-pro}
  \begin{tabularx}{\textwidth}{lccccc}
    \toprule
    \textbf{Dataset (Metric)} & \textbf{Full-Rank} & \textbf{PHLoRA (r32)} & \textbf{PHLoRA (r64)} & \textbf{PHLoRA (r512)} \\
    \midrule
    TAT--QA (Accuracy / EM)       & \textbf{89.38} / \textbf{62.48} & 87.96 / 55.22 (42) & 89.0 / 58.0 (50) & 89.0 / 61.0 (75) \\
    MKFE (Key Overlap / Value Overlap) & 99.50 / 24.75 & \textbf{100.0} / 24.00 (36) & \textbf{100.0} / \textbf{25.0} (43) & 50.0 / 23.0 (68) \\
    MedMCQA (Accuracy / F1)       & 69.4 / 69.42 & \textbf{70.0} / \textbf{70.0} (35) & \textbf{70.0} / \textbf{70.0} (41) & \textbf{70.0} / 69.70 (73) \\
    VQA--RAD (Similarity)           & 56.20        & \textbf{57.58} (41)        & 56.92 (50)                 & 57.07 (77)        \\
    CaptionGen (ROUGE\_L / CIDEr)  & \textbf{48.94} / 1.37       & 48.55 / 1.45 (35) & 48.85 / \textbf{1.48} (42)                 & 48.87 / 1.39 (70) \\
    \bottomrule
  \end{tabularx}
\end{table*}

\subsection{Ablation: Rank and Energy Preservation}
\label{sec:ablation}

We vary the PHLoRA rank (from 32 to 512) and report preserved energy $E_r$ (as defined in Equation\ref{eq:energy}). The results across all three Nova model sizes (Micro, Lite, Pro) are presented in Tables~\ref{tab:ablation-micro},~\ref{tab:ablation-lite}, and~\ref{tab:ablation-pro}, where each table reports test-set scores alongside preserved energy ($E_r$, \%) for different PHLoRA ranks and the full-rank reference.

Across all three Nova model scales (Micro, Lite, Pro), PHLoRA rank shows a clear correlation between preserved energy ($E_r$) and downstream task performance. Higher ranks consistently recover full-rank accuracy, while lower ranks maintain strong results with considerable efficiency gains. For Nova Micro (text-only), performance is stable across ranks, with r512 closely matching or slightly exceeding full-rank metrics on MedMCQA. In Nova Lite (multimodal), intermediate ranks such as r64 achieve performance comparable to or better than full-rank on tasks like VQA-RAD and CaptionGen. Similarly, in Nova Pro, r32 and r64 occasionally surpass full-rank scores, particularly in multimodal settings, though MKFE value overlap metrics appear more sensitive to rank and do not always improve with higher $E_r$. Overall, higher PHLoRA ranks reliably recover accuracy, while intermediate ranks can offer a strong balance between efficiency and performance across different model sizes and tasks.

\section{Conclusion}

We presented PHLoRA, a practical post-hoc method for deriving LoRA-compatible adapters directly from fully fine-tuned models, without requiring access to training data or gradients. Our experiments focused on three modalities—text, image, and video—using three Amazon Nova~\citep{novatechreport} models and five moderate-sized benchmarks, all in the supervised fine-tuning (SFT) setting. PHLoRA maintains competitive task accuracy while reducing inference GPU-hour costs by up to 4-fold compared to merged adapter inference, and by a similar or greater margin compared to full-rank model inference, in dynamic multi-adapter routing scenarios such as S-LoRA. This cost reduction reflects improvements in inference throughput, i.e., the number of tokens or requests processed per unit time, as demonstrated in S-LoRA~\citep{bing2024slora}.

PHLoRA provides a practical path for making all existing full-rank checkpoints adapter-ready, democratizing scalable inference for legacy models. 

\section{Future Work}

Several avenues remain for future research:
\begin{itemize}
    \item \textbf{Scaling to Larger and More Diverse Tasks:} Our current experiments are limited to moderate-sized SFT datasets. Future work should evaluate PHLoRA on larger-scale, more challenging benchmarks and additional modalities.
    \item \textbf{Advanced Tuning Strategies:} Extending PHLoRA to support advanced fine-tuning techniques such as DPO, PPO, or reward-based learning.
    \item \textbf{Extending Beyond Linear Layers:} While LoRA has been generalized to convolutions~\citep{zhong2024convolutionmeetsloraparameter}, post-hoc SVD-based extraction for higher-order tensors requires further research, potentially leveraging advanced tensor decompositions~\citep{kolda2009tensor} or alternative adapter parametrizations~\citep{chen2023parameterefficientfinetuningdesignspaces}.
    \item \textbf{Rank Selection and Usability:} Further developing practical methods for adaptive, data-free, or black-box rank selection, and enabling adapter extraction even when the base model is unavailable.
\end{itemize}

\section*{Limitations}

While \textbf{PHLoRA} offers a simple and effective post-hoc mechanism for adapter extraction, it comes with several important limitations.

PHLoRA is currently designed for standard linear (matrix-shaped) layers, as it relies on singular value decomposition (SVD) to extract low-rank adapters from weight differences. While LoRA and similar adapters have been extended to convolutional layers — either via kernel reshaping or structured convolutional approximations (e.g., \citep{zhong2024convolutionmeetsloraparameter}), post-hoc SVD extraction for convolutions or other higher-order tensors is non-trivial and depends on the decomposition or flattening strategy, which may lose spatial structure or interpretability. More generally, advanced tensor decompositions \citep{kolda2009tensor} or alternative adapter parametrizations \citep{chen2023parameterefficientfinetuningdesignspaces} would be required for such modules, which we leave to future work. Note also that attention “caches” refer to runtime data, not persistent parameters, and so are out of scope for PHLoRA.

In this work, we fix the adapter rank $r$ globally for all layers. Although we provide code to analyze energy-based rank selection, adaptive or per-layer rank scheduling—which could further improve the efficiency/accuracy tradeoff—remains for future work. Furthermore, while the preserved energy metric ($E_r$) is a useful indicator of information retention at a given rank, model quality on the target task does not always correlate perfectly with energy preservation. Thus, optimal adapter rank cannot be reliably selected solely from energy curves; empirical evaluation remains necessary.

PHLoRA assumes access to both base and fully fine-tuned weights. In settings where only the fine-tuned model is available (e.g., closed-source vendors), post-hoc adapter extraction is not directly possible.

The principal benefits of PHLoRA are realized in dynamic inference scenarios (e.g., S-LoRA or multi-adapter routing), where multiple adapters are loaded or swapped at runtime. In conventional merged-inference pipelines—where a single adapter is fused into the model for all requests—the practical advantage of post-hoc extraction is diminished, as cost and latency resemble standard LoRA or full-rank fine-tuning.

Our evaluation is limited to a set of public text, image, and video benchmarks. Results may differ for larger, more diverse real-world applications. While PHLoRA enables substantial inference cost reductions with dynamic adapter routing, there remains a modest runtime latency penalty versus full-rank merging; practical savings will depend on system-level batch sizes and workload characteristics.

\vspace{0.7em}
\noindent
We encourage future work to address these limitations by extending PHLoRA to non-linear modules, developing robust energy-aware or data-free rank selection strategies, enabling black-box or partial-weight extraction, and improving dynamic adapter composition schemes.

\bibliography{main}

\appendix

\section{Dataset Descriptions}
\label{app:datasets}

We provide summary statistics and descriptions for each benchmark used in this study. Scripts to reproduce the down-sampled and converted datasets will also be made available.

\begin{itemize}
    \item \textbf{TAT-QA}~\citep{zhu2021tatqa}: A table-augmented question answering dataset in the financial domain, requiring models to reason over both natural language and tabular data. \textit{Train:} 2,830; \textit{Test:} 565. License: MIT. Evaluated using Accuracy and Exact Match.
    
    \item \textbf{MKFE}~\citep{mkfe_hf}: Medical Knowledge from Extracts. Evaluates the ability to extract structured key-value medical facts from unstructured text. \textit{Key Overlap} measures the proportion of gold-standard keys correctly predicted; \textit{Value Overlap} measures the fraction of correct values among matched keys. \textit{Train:} 1,000; \textit{Test:} 200. License: Apache 2.0.
    
    \item \textbf{MedMCQA}~\citep{pmlr-v174-pal22a}: A large-scale medical multiple-choice question answering dataset. \textit{Train:} 20,000; \textit{Test:} 3,683. License: MIT. Evaluated using Accuracy and F1.
    
    \item \textbf{VQA-RAD}~\citep{lau2018rad}: Visual question answering over radiology images, requiring both visual and textual understanding. \textit{Train:} 1,793; \textit{Test:} 451. License: CC0 1.0 Universal. Evaluated by average normalized similarity.
    
    \item \textbf{CaptionGen}: A video captioning benchmark with 2,000 training and 500 test examples. Videos are sourced from MSVD~\citep{chen2011collecting}; captions are from the Multi-Source Video Captioning dataset~\citep{damo_nlp_sg_multi_source_video_captioning}. License: MIT. Evaluated using ROUGE\_L and CIDEr.
\end{itemize}

\vspace{1em}

\section{Implementation Details}

\textbf{Hardware.} All experiments were performed on AWS P5.48xlarge instances, each equipped with 8$\times$NVIDIA A100 80GB GPUs. Posthoc LoRA adapter extraction and energy analysis steps were also executed on the same hardware.

\textbf{Fine-tuning Hyperparameters.} We used the AdamW optimizer ($\beta_1{=}0.9$, $\beta_2{=}0.999$), a learning rate of $1 \times 10^{-5}$, batch size 32, and trained for 2 epochs.

\textbf{LoRA+ Training Hyperparameters.} 
\begin{itemize}
    \item \textbf{Nova Micro:} Learning rate $1 \times 10^{-5}$, \texttt{loraplus\_lr\_ratio} $16.0$, rank $r=32$, $\alpha=128$, \texttt{lora\_dropout} $0.01$, \texttt{target\_modules} = \texttt{[attention\_qkv, attention\_dense, mlp\_fc1, mlp\_fc2]}.
    \item \textbf{Nova Lite/Pro:} Learning rate $1 \times 10^{-5}$, \texttt{loraplus\_lr\_ratio} $8.0$, rank $r=32$, $\alpha=32$, \texttt{lora\_dropout} $0.01$, \texttt{target\_modules} = \texttt{[attention\_qkv, attention\_dense, mlp\_fc1, mlp\_fc2]}.
\end{itemize}

\textbf{PHLoRA Extraction.} SVD was performed per linear layer using PyTorch’s \texttt{torch.linalg.svd}. The default low-rank approximation used rank $r=32$, with ablations at ranks $r=64$ and $r=512$.

\textbf{Energy Plots.} Energy preserved at rank $r$, $E_r$, was calculated as in Equation.~\ref{eq:energy}. Plotting scripts are available at \texttt{scripts/plot\_energy.py}.

\vspace{1em}

\section{Reproducibility Checklist}
\begin{itemize}
  \item \textbf{Code:} All code—including SVD extraction, energy calculation, and evaluation scripts will be released at - \footnote{github URL to be added}.
  \item \textbf{Hyper-parameters:} Full grids in \texttt{configs/}.
  \item \textbf{Random seeds:} Fixed to 42.
\end{itemize}

\section{Optimality of SVD for Low-Rank Adapter Extraction}
\label{app:svd-optimality}

Given any real matrix $\Delta W \in \mathbb{R}^{m \times n}$, the Eckart–Young–Mirsky theorem~\citep{Eckart1936TheAO} states that the rank-$r$ matrix $\hat{W}_r = U_r \Sigma_r V_r^\top$ (where $U_r, \Sigma_r, V_r$ are the top $r$ components from the SVD of $\Delta W$) uniquely minimizes the Frobenius norm $\|\Delta W - \hat{W}_r\|_F$ over all matrices of rank at most $r$.

To see this, let $\Delta W = U \Sigma V^\top$ be the full SVD, with singular values $\sigma_1 \ge \sigma_2 \ge \cdots \ge \sigma_{\min(m,n)}$. The truncated approximation is
\[
\hat{W}_r = \sum_{i=1}^r \sigma_i u_i v_i^\top,
\]
and satisfies
\[
\|\Delta W - \hat{W}_r\|_F^2 = \sum_{i=r+1}^{\min(m,n)} \sigma_i^2.
\]
Therefore, by setting $A = U_r \operatorname{diag}(\sqrt{\Sigma_r})$ and $B = \operatorname{diag}(\sqrt{\Sigma_r}) V_r^\top$, as in PHLoRA, $AB = \hat{W}_r$ is the best rank-$r$ LoRA update (minimizing Frobenius error).

For more details, see~\citep{Eckart1936TheAO,golub2013matrix}.

\end{document}